\documentclass[letterpaper]{article}

\usepackage{natbib,alifeconf}  
\usepackage{subfigure}
\usepackage{graphicx}
\usepackage{multicol}
\usepackage{setspace}
\usepackage{enumitem}
\usepackage{hyperref} 

\title{ALIFE2023 template}

\title{EINCASM: Emergent Intelligence in Neural Cellular Automaton Slime Molds}
\author{Aidan Barbieux \and Rodrigo Canaan\\
\mbox{}\\
 California Polytechnic State University, San Luis Obispo, CA 93410 \\
 abarbieu@calpoly.edu} 

\begin{document}
\maketitle

\begin{abstract}
\vspace{-\baselineskip}

This paper presents EINCASM, a prototype system employing a novel framework for studying emergent intelligence in organisms resembling slime molds. EINCASM evolves neural cellular automata with NEAT to maximize cell growth constrained by nutrient and energy costs. These organisms capitalize physically simulated fluid to transport nutrients and chemical-like signals to orchestrate growth and adaptation to complex, changing environments. Our framework builds the foundation for studying how the presence of puzzles, physics, communication, competition and dynamic open-ended environments contribute to the emergence of intelligent behavior. We propose preliminary tests for intelligence in such organisms and suggest future work for more powerful systems employing EINCASM to better understand intelligence in distributed dynamical systems.

\end{abstract}

\vspace{-\baselineskip}
\vspace{-\baselineskip}

\section{Introduction}


Emergent intelligence is a phenomenon where the interaction of simple components combine to produce novel behavior aimed at achieving a goal. Consider, for example, how humans distributed across a landscape form powerful societies or how the combination of many simple cells results in organisms that solve puzzles to acquire food. This property begins to manifest in complex particle systems \citep{schmicklHowLifelikeSystem2016, gregorSelfOrganizingIntelligentMatter2021}, although the resultant behavior can be limited and challenging to interpret. By introducing a well defined goal and computer vision techniques, neural cellular automata (NCA) can produce impressive self-organization and intelligence, such as morphogenesis, maze solving, and number recognition \citep{nicheleCANEATEvolvedCompositional2018, endoNeuralCellularMaze, mordvintsevGrowingIsotropicNeural2022, randazzoSelfclassifyingMNISTDigits2020}, however, these systems are not open ended or lifelike. We propose EINCASM as a bridge between these systems to produce life-like organisms with interpretable intelligence that display emergence. By focusing on the simple physiology of slime molds, which have well-studied intelligent foraging behavior, our framework aims to be tractable and testable. A main novelty of our work is the inclusion of physical constraints and fluid simulation which organisms must orchestrate via chemical-like communication to grow and adapt.

\begin{figure*}
    \includegraphics[width=\linewidth]{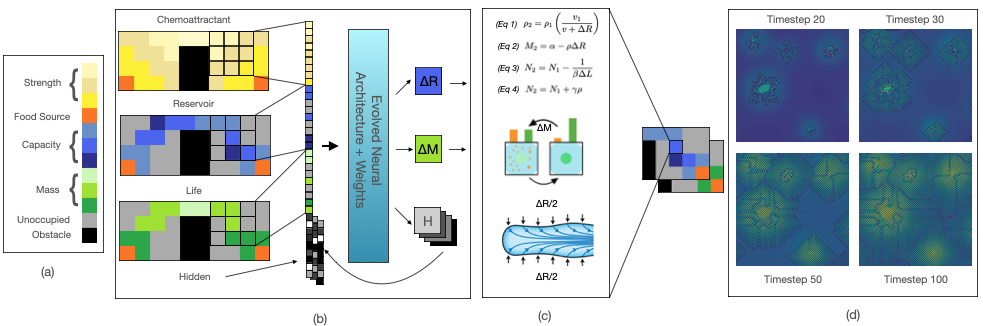}
    \caption{(a): The channels of the environment perceived by the NCA. (b) The application of the evolved neural architecture. (c) The enforcement of physical constraints and interface with physical simulation. (d) An example organism's reservoir and chemoattractant as it creates a rudimentary transport network around an obstacle. Equation 1 relates the cytoplasm pressure $\rho$ to the reservoir of size $v$ and the change in reservoir size $\Delta R$. Equation 2 describes the cost in cell mass $M$ for a change in reservoir size with a movement cost of $\alpha$. Equation 3 represents the cost in nutrient $N$ for growth in cell mass $\Delta M$ with a growth cost of $\beta$. Finally, equation 4 defines nutrient uptake at a rate of $\gamma$ on a food source channel $F$.}
    \label{fig:sys}
\end{figure*}

\section{Evolution and Intelligence Tests}

To maintain homeostasis against the increase of entropy, organisms must find and use energy from the environment. Other organisms and events, such as weather, affect the environment and require adaptation. EINCASM explores these adaptations by evolving NCA that use limited nutrients to replicate and move in a dynamic and complex environment, leading to intelligent behavior to acquire nutrients, such as pathfinding (figure \ref{fig:sys}d). This extends the work of \cite{nicheleCANEATEvolvedCompositional2018} and \cite{mordvintsevGrowingIsotropicNeural2022} on NCA by leaving the ideal behavior of the system undefined. Instead, fitness is determined by the ability of the NCA to accomplish a goal, i.e., growth, in a wide range of circumstances -- which is a common definition of intelligence. 

Following \cite{nicheleCANEATEvolvedCompositional2018, stanleyDesigningNeuralNetworks2019}, we use a Compositional Pattern Producing Network (CPPN) as the ruleset for the NCA (figure \ref{fig:sys}b). This represents the cellular physiology of the organism. To evolve the CPPN we simulate a variable-length lifecycle where a NCA grows given physical constraints. Throughout this lifecycle, nutrient sources are removed, cells are degraded, and obstacles move. This is analagous to, for example, a branch falling on a section of slime mold. At the end of the time period fitness is determined simply as the sum of cell mass.

In the current implementation of EINCASM, single organisms are grown in contained, simple environments. However, future systems could include multiple organisms competing for resources with no set notion of an individual, as exemplified by \cite{gregorSelfOrganizingIntelligentMatter2021} and suggested by \cite{sorosIdentifyingNecessaryConditions2014} to promote truly open-ended complexity.

To measure the intelligence of the organisms produced by EINCASM-like systems, we propose biologically inspired tests. Each test consists of introducing an organism to a novel ``puzzle" environments where certain adaptations are required to thrive. By measuring completion, speed, and growth rates, a sort of IQ can be recorded. The preliminary tests are proposed:

\begin{itemize}[itemsep=0mm, parsep=0pt]
    \item \textbf{Coordination}: Will the organism redistribute its mass to explore new areas when a nutrient source is removed?
    \item \textbf{Pathfinding}: Can the organism solve a complex maze to reach rich nutrients?
    \item \textbf{Learning and Knowledge sharing}: Given a repeated feature (e.g. deceptive chemoattractant), does the organism adopt new behavior? Can this behavior be shared with part of the organism that hasn't seen this deception? \citep{vogelDirectTransferLearned2016}
    \item \textbf{Adversarial Nutrient Storage}: Can an organism protect nutrients from competing species and recover them in adverse circumstances?
\end{itemize}

\section{Prototype System}

EINCASM organisms are defined as a set of cells, each of which performs a local operation and collaborates with other cells either directly by sharing a neighborhood or indirectly by directing cytoplasmic flow via reservoir contraction. Each cell has identical physiology, but is differentiated by its local environment and the signals from other cells, similar to stem cells forming different parts of an organism during development.

Our  system is represented by a square tile grid of real-valued static, dynamic, and hidden channels (figure~\ref{fig:sys}a). Static channels define the environment with obstacles, poison, food sources, and chemoattractant, inspired by \citet{tsompanasEvolvingTransportNetworks2015}. Contents of the hidden channels can be used by the agent and are interpreted by \citet{mordvintsevGrowingIsotropicNeural2022} as chemical or electrical signaling between cells.

The main novelty of our approach lies in the dynamic channels, which represent cell mass, reservoir size, and nutrient containing cytoplasm. These are determined jointly by the agent and physical simulation. This approach was explored by~\citet{gregorSelfOrganizingIntelligentMatter2021} but, as far as we are aware of, is novel in the context of NCA.

On each time step, cells are updated stochastically by applying an evolved 3x3 convolutional kernel to produce new hidden channels (updated without modification), and a desired change in reservoir size and cell mass. Figure \ref{fig:sys}c describes how cell mass and reservoir size are physically constrained by nutrient consumption and cell mass can be converted freely into nutrients to support network adaptation.

The final reservoir size is then used to update cytoplasmic flow using the Lattice Boltzmann method, following \citep{conningtonPeristalticLBM2009}. Briefly, this enables the organism to produce peristaltic-like motion which pumps nutrients along adjacent sequences of reservoirs. This effect is preferential towards parallel sets of large reservoirs and is important to slime mold's motility and adaptive behavior \citep{lewisInvestigationPeristalticPumping2016, rayInformationTransferFood2019}.

\section{Preliminary Results and Future Work}

While a thorough quantitative analysis is still required, our current prototype already demonstrates preliminary signs of coordination and pathfinding within handcrafted environments comprising obstacles and poison. An example can be seen in Figure~\ref{fig:sys}d. Our physical simulation is limited to directly modifying pressure at small scales relative to reservoir size rather than dynamic fluid boundary conditions.

In the future, we plan to enhance the scale and realism of fluid simulation in addition to evaluating competition in multi-agent settings, dynamically generated environments, linguistic evolution (utilizing the hidden channel for pattern communication), and multi-scale competency or modularity.



\footnotesize
\bibliographystyle{apalike}
\bibliography{ENCASM-ALife2023-Bib} 

\end{document}